%% file: main.tex
\documentclass{article}
\usepackage{ijcai22}

\usepackage{times}
\usepackage{soul}
\usepackage{url}
\usepackage[hidelinks]{hyperref}
\usepackage[utf8]{inputenc}
\usepackage[small]{caption}
\usepackage{graphicx}
\usepackage{amsmath}
\usepackage{amsthm}
\usepackage{booktabs}
\usepackage{algorithm}
\title{IJCAI--22 Formatting Instructions}

\author{
    Author Name
    \affiliations
    Affiliation
    \emails
    pcchair@ijcai-22.org
}

\author{
Zichuan Lin \thanks{Equal contribution} \quad
Junyou Li \footnotemark[1] \quad
Jianing Shi \footnotemark[1] \quad
Deheng Ye \quad
Qiang Fu \quad
Wei Yang 
\affiliations
Tencent AI Lab, Shenzhen, China\\
\emails
\{zichuanlin, junyouli, jianingshi, dericye, leonfu, willyang\}@tencent.com
}

\usepackage{natbib}
\usepackage[english]{babel}
\usepackage[utf8]{inputenc} %
\usepackage[T1]{fontenc}    %
\usepackage{hyperref}       %
\usepackage{url}            %
\usepackage{booktabs}       %
\usepackage{amsfonts}       %
\usepackage{nicefrac}       %
\usepackage{microtype}      %
\usepackage{amsmath,amssymb}
\usepackage{mathtools}
\usepackage{xcolor}
\usepackage{enumerate}
\usepackage{booktabs}
\usepackage{comment}
\usepackage{algorithm,algorithmicx}
\usepackage[noend]{algpseudocode}
\usepackage{bbm}
\usepackage{wrapfig}
\usepackage{subfigure}
\usepackage{multirow}
\usepackage{diagbox}
\usepackage{enumitem}

\def\shownotes{1}  %
\ifnum\shownotes=1
\newcommand{\authnote}[2]{[#1: #2]}
\else
\newcommand{\authnote}[2]{}
\fi

\newcommand{\E}{\mathbb{E}}
\newcommand{\Esub}[1]{\underset{#1}{\E}}

\newcommand{\cL}{\mathcal{L}}
\newcommand{\cO}{\mathcal{O}}
\newcommand{\cD}{\mathcal{D}}
\newcommand{\cB}{\mathcal{B}}
\newcommand{\cS}{\mathcal{S}}

\newcommand{\cA}{\mathcal{A}}

\newcommand{\bbR}{\mathbb{R}}
\DeclareMathOperator*{\argmax}{arg\,max}

\renewcommand{\paragraph}[1]{{\noindent {\bf #1.}}}

\title{JueWu-MC: Playing Minecraft with Sample-efficient \\Hierarchical Reinforcement Learning}

\begin{document}
\maketitle

\begin{abstract}

Learning rational behaviors in open-world games like Minecraft remains to be challenging for Reinforcement Learning (RL) research due to 
the compound challenge of partial observability, high-dimensional visual perception and delayed reward. 
To address this, we propose JueWu-MC, a sample-efficient hierarchical RL approach equipped with representation learning and imitation learning to deal with perception and exploration. 
Specifically, our approach includes two levels of hierarchy, where the high-level controller learns a policy to control over options and the low-level workers learn to solve each sub-task.
To boost the learning of sub-tasks, we propose a combination of techniques including 1) action-aware representation learning which captures underlying relations between action and representation, 2) discriminator-based self-imitation learning for efficient exploration, and 3) ensemble behavior cloning with consistency filtering for policy robustness. 
Extensive experiments show that JueWu-MC significantly improves sample efficiency and outperforms a set of baselines by a large margin.
Notably, we won the championship of the NeurIPS MineRL 2021 research competition and achieved the highest performance score ever.

\end{abstract}

\input{intro}

\input{relatedwork}

\input{method}

\input{exp}

\input{conclusion}

\newpage
\newpage

\bibliographystyle{named}
\bibliography{paper_used}

\newpage
\input{appendix}

\end{document}

%% file: intro.tex
\section{Introduction}

Deep reinforcement learning (DRL) has shown great success in many genres of games, including board game~\citep{silver2016mastering}, Atari~\citep{mnih2013playing}, simple first-person-shooter (FPS)~\citep{huang2019combo}, real-time strategy (RTS)~\citep{vinyals2019grandmaster}, multiplayer online battle arena (MOBA)~\citep{berner2019dota}, etc. 
Recently, open-world games have been attracting attention due to its playing mechanism and similarity to real-world control tasks~\citep{guss2021minerl}. 
Minecraft, as a typical open-world game, has been increasingly explored for the past few years~\citep{oh2016control,tessler2017deep,MineRLDataset,kanervisto2020playing,skrynnik2021forgetful,mao2021seihai}. 

Compared to other games, the characteristics of Minecraft make it a suitable testbed for RL research, as it emphasizes exploration, perception and construction in a 3D open world \citep{oh2016control}. 
The agent is only provided with partial observability and occlusions. 
The tasks in the game are chained and long-term. 
Generally, human can make rational decisions to explore basic items and construct desired higher-level items using a reasonable amount of samples, while it can be hard for an AI agent to do so autonomously. 
Therefore, to facilitate the efficient decision-making of agents in playing Minecraft,  MineRL~\citep{MineRLDataset} has been developed as a research competition platform, which provides human demonstrations and encourages the development of sample-efficient RL agents for playing Minecraft.
Since the release of MineRL, a number of efforts have been made on developing Minecraft AI agents, e.g., ForgER~\citep{skrynnik2021forgetful}, SEIHAI~\citep{mao2021seihai}.

However, it is still difficult for existing RL algorithms to mine items in Minecraft due to the compound challenge it poses, expanded below. 

\paragraph{Long-time Horizons}
In order to achieve goals (e.g., mining a diamond) in Minecraft, the agent is required to finish a variety of sub-tasks (e.g., log, craft) that highly depend on each other. Due to the sparse reward, it is hard for agents to learn long-horizon decisions efficiently. Hierarchical RL from demonstrations~\citep{le2018hierarchical,pertsch2020accelerating} has been explored to leverage the task structure to accelerate the learning process. However, learning from unstructured demonstrations without domain knowledge remains challenging. %

\paragraph{High-dimensional Visual Perception}
Minecraft is a flexible 3D first-person game revolving around gathering resources (i.e., explore) and creating structures and items (i.e., construct). In this environment, agents are required to deal with high-dimensional visual input to enable efficient control. However, agent's surroundings are varied and dynamic, which poses difficulties to learning a good representation. 

\paragraph{Inefficient Exploration}
With partial observability, the agent needs to explore in the right way and collect information from the environment so as to achieve goals. A naive exploration strategy can waste a lot of samples on useless exploration. Self-imitation Learning (SIL)~\citep{oh2018self} is a simple method that learns to reproduce past good behaviors to incentivize deep exploration. However, SIL is not sample-efficient because its advantage-clipping operation causes a waste of samples. Moreover, SIL does not make use of the transitions between samples.

\paragraph{Imperfect Demonstrations}
Human demonstrations in playing Minecraft are highly distributional diverse \citep{kanervisto2020playing}. 
Also, there exists noisy data due to the imperfection of human operation \citep{MineRLDataset}. 

To address the aforementioned compound challenges, we develop an efficient hierarchical RL approach equipped with novel representation and imitation learning techniques. Our method makes effective use of human demonstrations to boost the learning of agents and enables the RL algorithm to learn rational behaviors with high sample efficiency.

\paragraph{Hierarchical Planing with Prior}
We first propose a hierarchical RL (HRL) framework with two levels of hierarchy, where the high-level controller automatically extracts sub-goals in long-horizon trajectories from the unstructured human demonstrations and learns a policy to control over options, while the low-level workers learn sub-tasks to achieve sub-goals by leveraging both demonstrations dispatched by the high-level controller and interactions with environments. Our approach automatically structures the demonstrations and learns a hierarchical agent, which enables better decision over long-horizon tasks. Under our HRL framework, we devise the following key techniques to boost agent learning. 

\paragraph{Action-aware Representation Learning}
Although some prior works~\citep{huang2019combo} proposed using auxiliary tasks (e.g., enemy detection) to better understand the 3D world, such methods require a large amount of labeled data. 
We propose a self-supervised action-aware representation learning (A2RL) technique, which learns to capture the underlying relations between action and representation in 3D visual environments like Minecraft. 
As we will show, A2RL not only enables effective control by learning a compact representation but also improves the interpretability of the learned policy.

\paragraph{Discriminator-based Self-imitation Learning}
As mentioned, existing self-imitation learning is advantage-based and becomes sample-inefficient for handling tasks in Minecraft, as it wastes a lot of samples due to the clipped objective and does not utilize transitions between samples. Therefore, we propose discriminator-based self-imitation learning (DSIL) which leverages self-generated experiences to learn self-correctable policies for better exploration. 

\paragraph{Ensemble Behavior Cloning with Consistency Filtering}
Learning a robust policy from imperfect demonstrations is difficult~\citep{wu2019imitation}. To address this issue, we first propose \textit{consistency filtering} to identify the most common human behavior, and then perform \textit{ensemble behavior cloning} to learn a robust agent with reduced uncertainty.

In summary, our contributions are: 1) We propose JueWu-MC, a sample-efficient hierarchical RL approach, equipped with action-aware representation learning, discriminator-based self-imitation, and ensemble behavior cloning with consistency filtering, for training Minecraft AI agents.  2) Our approach outperforms competitive baselines by a significantly large margin and achieves the best performance ever throughout the MineRL competition history. Thorough ablations and visualizations are further conducted to help understand why our approach works.

%% file: relatedwork.tex
\section{Related Work}

\paragraph{Game AI}
Game has long been a preferable field for artificial intelligence research. 
AlphaGo~\citep{silver2016mastering} mastered the game of Go with DRL and tree search. 
Since then, DRL has been used in other more sophisticated games, including StarCraft (RTS)~\citep{vinyals2019grandmaster}, Google Football (Sports)~\citep{kurach2020google}, VizDoom (FPS)~\citep{huang2019combo}, Dota (MOBA)~\citep{berner2019dota}. %
Recently, the 3D open-world game Minecraft %
is drawing rising attention. 
\cite{oh2016control} showed that existing RL algorithms suffer from generalization in Minecraft and proposed a new memory-based DRL architecture. 
\cite{tessler2017deep} proposed H-DRLN, a combination of a deep skill array and a skill distillation system, to promote lifelong learning and transfer knowledge among different tasks in Minecraft. 
Since MineRL was held in 2019, many solutions have been proposed to learn to play in Minecraft. %
There works can be grouped into two categories: 1) end-to-end learning~\citep{amiranashvili2020scaling,kanervisto2020playing,scheller2020sample}; 2) HRL with human demonstrations~\citep{skrynnik2021forgetful,mao2021seihai}. 
Our approach belongs to the second category.
In this category, prior works leverage the structure of the tasks and learn a hierarchical agent to play in Minecraft --- ForgER~\citep{skrynnik2021forgetful} proposed a hierarchical method with forgetful experience replay to allow the agent to learn from low-quality demonstrations; \cite{mao2021seihai} proposed SEIHAI that fully takes advantage of the human demonstrations and the task structure.

\paragraph{Sample-efficient Reinforcement Learning}
Our work is to build a sample-efficient RL agent for playing Minecraft, and we thereby develop a combination of efficient learning techniques. We discuss the most relevant works below. 

Our work is related to recent HRL research that builds upon human priors. 
To expand, \cite{le2018hierarchical} proposed to warm-up the hierarchical agent from demonstrations and fine-tune with RL algorithms. \cite{pertsch2020accelerating} proposed to learn a skill prior from demonstrations to accelerate HRL algorithms. 
Compared to existing works, we are faced with the highly unstructured demo in 3D first-person video games played by the crowds.
We address this challenge by structuring the demonstrations and defining sub-tasks and sub-goals automatically.

Representation learning in RL has two broad directions: self-supervised learning and contrastive learning. The former~\citep{wu2021self} aims at learning rich representations for high-dimensional unlabeled data to be useful across tasks, while the latter~\citep{srinivas2020curl} learns representations that obey similarity constraints in a dataset organized by similar and dissimilar pairs.
Our work proposes a novel self-supervised representation learning method that can measure action effects in 3D video games.

Existing methods use curiosity or uncertainty as a signal for exploration~\citep{pathak2017curiosity,burda2018exploration} so that the learned agent is able to cover a large state space.
However, the exploration-exploitation dilemma, given the sample efficiency consideration, drives us to develop self-imitation learning (SIL)~\citep{oh2018self} methods that focus on exploiting past good experiences for better exploration. 
Hence, we propose discriminator-based self-imitation learning (DSIL) for efficient exploration.

Our work is also related to learning from imperfect demonstrations, such as DQfD~\citep{hester2018deep} and Q-filter~\citep{nair2018overcoming}. %
Most methods in this field leverage online interactions with the environment to handle the noise in demonstrations. 
We propose ensemble behavior cloning with consistency filtering (EBC) which leverages imperfect demonstrations to learn robust policies in playing Minecraft. %

%% file: method.tex
\section{Method}

In this section, we first introduce our overall HRL framework, and then illustrates the details of each component. %

\subsection{Overview}

\begin{figure}[t]
	\centering
	\includegraphics[width=0.46\textwidth]{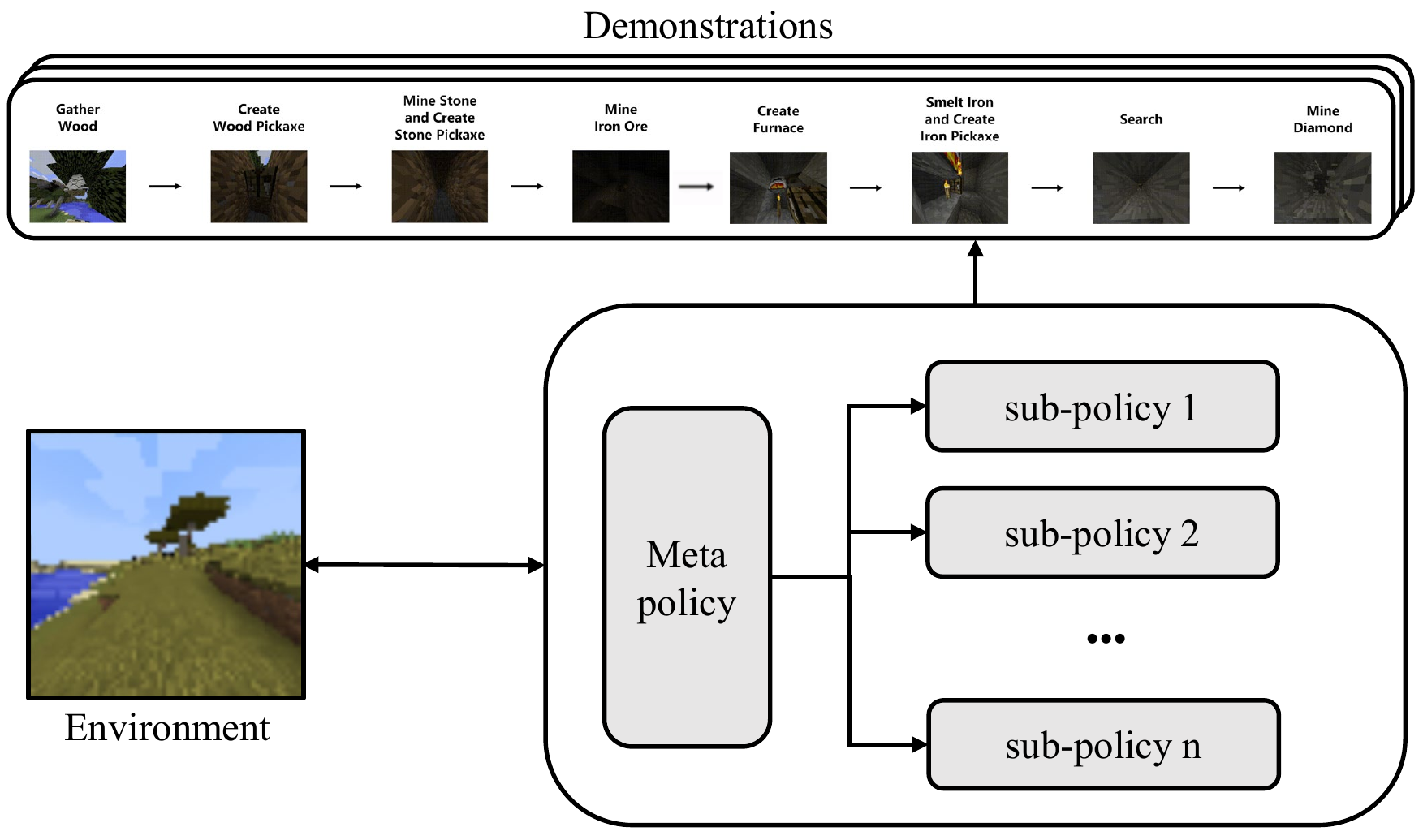}
	\caption{The overall framework of our method. }
	\label{fig:framework}
	\vspace{-2mm}
\end{figure}

Figure~\ref{fig:framework} shows our overall framework. 
We define the human demonstrations as $\cD = \{\tau_0, \tau_1, \tau_2, ...\}$ where $\tau_i$ represents a long-horizon trajectory containing states, actions and rewards. 
The provided demonstrations are unstructured in that there are no explicit signals to specify sub-tasks and sub-goals.

We first define \textit{atomic skill} as an individual skill that gets a non-zero reward. %
Then, we define sub-tasks and sub-goals based on \textit{atomic skill}. 
To define reasonable sub-tasks, we examine the degree of reward delay for each atomic skill. We keep those atomic skills with long reward delay as individual sub-tasks because they require executing a long sequence of actions to achieve a delayed reward. Meanwhile, we merge those adjacent atomic skills with short reward delay into one sub-task. 
By doing so, we get $n$ sub-tasks (a.k.a stages) in total. 
To define a sub-goal for each sub-task, we extract the most common human behavior pattern and use the last state in each sub-task as its sub-goal.
In this way, we get \textit{structured demonstrations} ($\cD \rightarrow \{\cD_0, \cD_1, ..., \cD_{n-1}\}$) with sub-tasks and sub-goals that are used to train the hierarchical agent. %
With the structured demonstrations, we train the meta-policy by imitation learning, and train the sub-policies to solve sub-tasks by leveraging both demonstrations and interactions with the environment, as described below.

\subsection{Meta- and Sub-policies}

\paragraph{Meta-policy}
We train a meta-policy that maps continuous states to discrete indices ($0,1,...,n-1$) that specifies which option to be used. Given state space $\bar{\cS}$ and discrete option space $\cO$, the meta-policy is defined as $\pi^m_{\rho} (o | \bar{s})$, where $\bar{s} \in \bar{\cS}$ is an inventory vector that summarizes the agent's collected items, $o \in \cO$ is a discrete value and $\rho$ represents parameters. $\pi^m_{\rho} (o | \bar{s})$ specifies the conditional distribution over the discrete options. 
To train the meta-policy, we generate training data $(\bar{s}, i)$ where $i$ represents the $i$-th stage and $\bar{s} \in \cD_i$ is sampled from the demonstrations of the $i$-th stage.
The meta-policy is trained using negative log-likelihood (NLL) loss:
\begin{equation}
\footnotesize
    \min_{\rho} \sum_{i=0}^{n-1} \Esub{\bar{s} \in \cD_i} \left[ - \log \pi^m_{\rho} (i | \bar{s}) \right].
\end{equation}
During inference, the meta-policy generates options by taking argmax on the distribution $\hat{o} = \argmax_o \pi^m_{\rho} (o | \bar{s}) $.

\paragraph{Sub-policy}
In Minecraft, sub-tasks can be grouped into two main types: gathering resources, and crafting items. In the first type (gathering resources), agents need to navigate and gather sparse rewards by observing high-dimensional visual inputs which are varied and dynamic. %
In the second type (crafting items), agents need to execute a sequence of actions robustly. %

In typical HRL, action space of the sub-policies is pre-defined according to prior knowledge. However, in the MineRL 2020\&2021, handcrafted action space is prohibited. Besides, action space is obfuscated in both human demonstrations and the environment. Directly learning in this continuous action space is challenging as exploration in a large continuous action space can be inefficient. Therefore, we use KMeans~\citep{krishna1999genetic} to cluster actions for each sub-task using demonstration $\cD_i$, and perform reinforcement learning and imitation learning based on the clustered action space.

In the following section, we describe how to learn sub-policies efficiently to solve these two kinds of sub-tasks.

\subsection{Learning Sub-policies to Gather Resources}
To efficiently solve this kind of sub-tasks, we propose action-aware representation learning as well as discriminator-based self-imitation learning to facilitate the learning process of sub-policies. We show the model architecture in Figure~\ref{fig:model1}. The full algorithm is shown in Appendix~\ref{app:algo}.

\begin{figure}[t]
	\centering
	\includegraphics[width=0.46\textwidth]{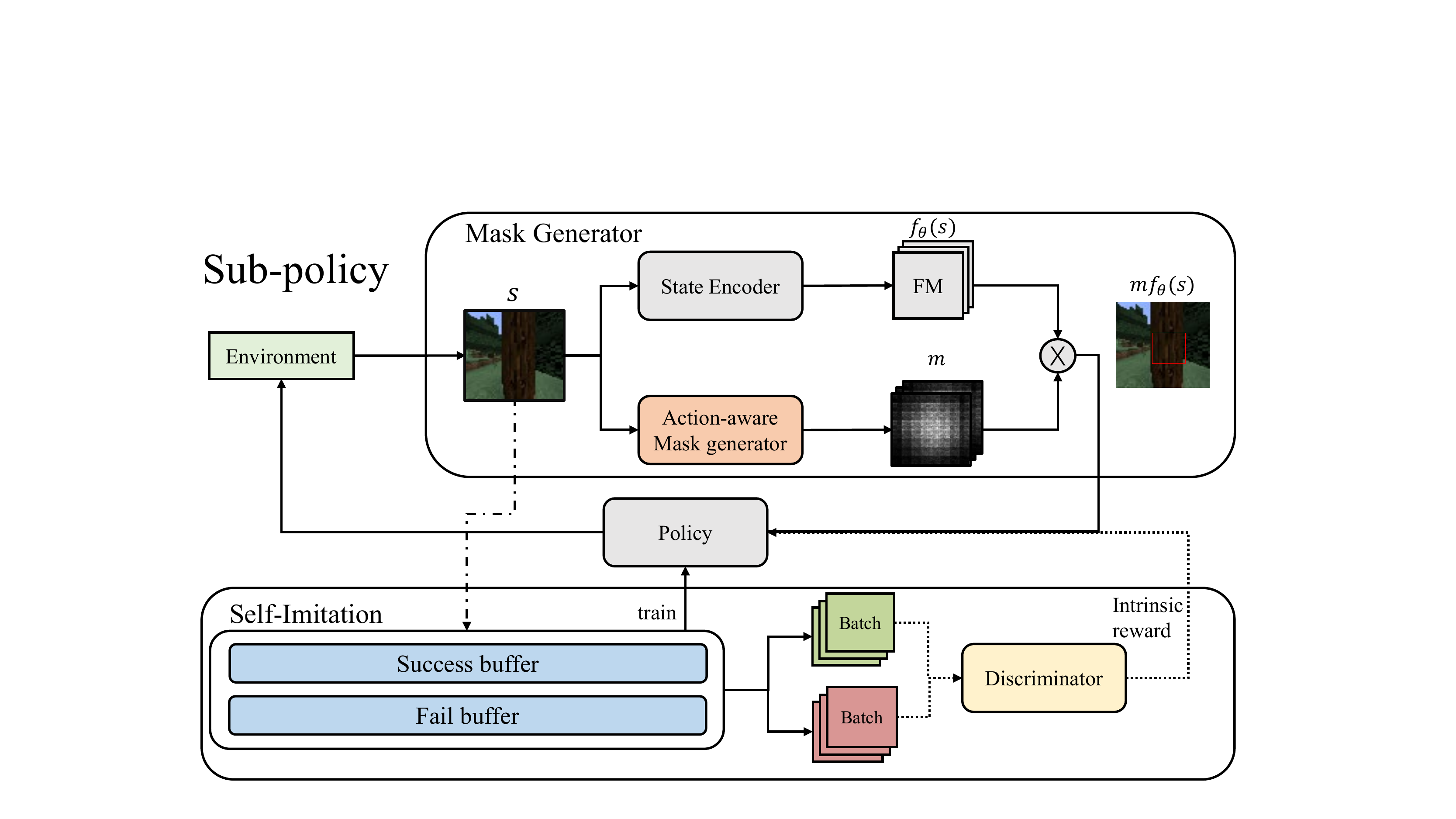}
	\caption{Model architecture for visual-based resources gathering sub-tasks. The architecture contains two main parts, i.e., a mask generator for action-aware representation and a self-imitation mechanism. The mask generator serves as attention to help agents focus on salient parts of the visual input. The self-imitation mechanism help agents quickly latch on good behaviors and drive deep exploration.}
	\label{fig:model1}
\end{figure}

\subsubsection{Action-aware Representation Learning} %

Learning compact representation is crucial to improve sample efficiency in reinforcement learning~\citep{lesort2018state}. 
To tackle the challenge of learning good representation in 3D open world, we start by observing that in first-person 3D environments, different actions have their own effects --- each action acts on a local part of the high-dimensional observations. 
For example, in Minecraft, the attack action aims to break and acquire the block in front of the agent, while the camera action aims to adjust the agent's camera perspective.
Motivated by this observation, we propose action-aware representation learning (A2RL), to learn representation that can capture the underlying relation with actions. %

To achieve so, we leverage the dynamic property from environments. 
Specifically, we learn a mask net on feature map for each action to capture dynamic information between the current and next states. %
Denote the feature map as $f_{\theta}(s) \in \bbR^{H \times W}$ and the mask net as $m_{\phi}(s, a) \in [0,1]^{H \times W}$, where $\theta$ and $\phi$ represent parameters of convolution neural network of the policy and mask net. 
Given a transition tuple $(s, a, s')$, the loss function for training the mask is as follows:
\begin{equation} \label{eq:mask_L1}
\scriptsize
    \cL_{m}^1(\phi) = 
    \Esub{s,a,s' \sim \cD}  \bigg[ \Big\|  g_{\psi_a}\big(  (1 - m_{\phi}(s, a)) \odot f_{\theta}(s) \big)  - f_{\theta}(s') \Big\|_2  \bigg],
\end{equation}
\begin{equation} \label{eq:mask_L2}
\scriptsize
    \cL_{m}^2(\phi) = 
    \Esub{s,a,s' \sim \cD}  \bigg[ - \Big\|  g_{\psi_a}\big(  m_{\phi}(s, a) \odot f_{\theta}(s) \big)  - f_{\theta}(s') \Big\|_2  \bigg],
\end{equation}
\begin{equation}\label{eq:mask}
\footnotesize
    \cL_{m}(\phi) = \cL_{m}^1(\phi) + \eta \cL_{m}^2(\phi),
\end{equation}
where $g_{\psi_a}$ is a linear projection function parameterized by learnable parameters $\psi_a$; %
$\odot$ represents element-wise product; %
$\eta$ is a hyper-parameter to trade off two objectives.

To optimize Eq~\ref{eq:mask}, we use a two-stage training process. In the first stage, we train the linear projection network $g_{\psi_a}$ using the following objective:
\begin{equation} \label{eq:g}
\footnotesize
    \cL_{g}(\psi_a) = \Esub{s,a,s' \sim \cD}  \bigg[  \Big\|  g_{\psi_a}\big( f_{\theta}(s) \big)  - f_{\theta}(s') \Big\|_2  \bigg].
\end{equation}
This objective learns to recover information of $s'$ from $s$ in latent space, which is equivalent to learning a dynamic model to predict next state given current state and action. Note that the parameter $\psi_a$ is dependent with action $a$. In the second stage, we fix the learned linear function $g_{\psi_a}$ and use Eq~\ref{eq:mask} to optimize the mask net.

Intuitively, on the one hand, by minimizing Eq~\ref{eq:mask_L1}, the mask net will learn to mask out the parts that introduce uncertainty to the model-based prediction, while remaining adequate information to predict the next state. On the other hand, by minimizing Eq~\ref{eq:mask_L2}, the mask net will tend to pay attention to as little information as possible, trying to introduce uncertainty to the prediction.
Therefore, by minimizing them jointly in Eq~\ref{eq:mask}, the mask net can learn to focus on \textit{local} parts of the current image that introduce uncertainty to the dynamic model. This is similar to human curiosity, which pays attention to the part that is uncertain to themselves. %

A2RL is reminiscent of dynamics-based representation~\cite{whitney2019dynamics}. However, dynamics-based representation learning aims to learn representation that is capable to imagine dynamics with a long horizon.
Our approach aims to learn the underlying relations between representation and action by leveraging one-step dynamic prediction --- this provides the agent with multi-view representations that reveal the effects of different actions. 
The learned representations can then be combined with any off-the-shelf RL algorithms to improve sample efficiency.

For policy-based methods, we plug our learned representations into policy $\pi_{\theta}(a | (1 + m_{\phi}(s, a)) \odot f_{\theta}(s) )$ for effective perception and efficient back-propagation of policy gradient.
For value-based methods, we combine our learned representation directly with Q-value functions $Q_{\theta_q}((1 + m_{\phi}(s, a)) \odot f_{\theta}(s),a)$. The learning of Q-value function can be done using any Q-learning based algorithms. %

\subsubsection{Discriminator-based Self-imitation Learning}

Self-imitation Learning (SIL)~\citep{oh2018self} is considered as a simple but effective way to solve hard-exploration tasks. 
SIL uses an advantage clipping technique to bias the agent towards good behaviors, which we call it as \textit{advantage-based self-imitation learning} (ASIL). However, %
SIL is not sample-efficient due to the clipping mechanism. Besides, SIL does not leverage the transition between samples. %

To address the issues of SIL, we propose \textit{discriminator-based self-imitation learning} (DSIL). %
Unlike ASIL, DSIL does not use advantage clipping. Our intuition is that the agent should be encouraged to visit the state distribution that is more likely to lead to goals. 

To do so, DSIL first learns a discriminator to distinguish between states from successful and failed trajectories (i.e., ``good'' and ``bad'' states), and then uses the learned discriminator to guide exploration. 
Specifically, We maintain two replay buffers $\cB_{i}^{+}$ and $\cB_{i}^{-}$ to store successful and failed trajectories respectively. 
During learning, we treat data from $\cB_{i}^{+}$ as positive samples and data from $\cB_{i}^{-}$ as negative samples to train the discriminator. Denote the discriminator as $D_{\xi}: \cS \rightarrow [0,1]$ which is parameterized by parameters $\xi$. We train the discriminator with the following objective:
\begin{equation} \label{eq:discriminator}
\footnotesize
    \max_{\xi} \Esub{s \sim \cB_{i}^{+}} \left[\log D_{\xi}(s) \right] + \Esub{s \sim \cB_{i}^{-}} \left[1 - \log D_{\xi}(s) \right].
\end{equation}
Intuitively, this objective encourages $D_{\xi}(s)$ to output high values for good states while giving low values for bad states. For those states that are not distinguishable, $D_{\xi}(s)$ tends to output 0.5. 
The learned discriminator captures the good state distribution that leads to goals and the bad state distribution that leads to failure.

We then use the trained discriminator to provide intrinsic rewards for policy learning to guide exploration. The intrinsic reward is defined as:
\begin{equation} \label{eq:intrinsic}
\scriptsize
    \bar{r}(s,a,s') = \begin{cases} +1,& D_{\xi}(s') > 1-\epsilon \\ -1,& D_{\xi}(s') < \epsilon \end{cases}
\end{equation}
where $\epsilon \in (0, 0.5)$ is a hyper-parameter to control the confidence interval of $D_{\xi}$. This reward drives the policy to explore in regions that previously get successful trajectories. By training the policy with this intrinsic reward, we encourage the policy to stay close to good state distribution so as to reproduce the agent's past good decisions. Also, it is worth noting that DSIL encourages the policy to be \textit{self-correctable} --- it helps push the agent to the good state distribution even when the agent falls into the bad state distribution accidentally.

\begin{table*}[t!]
\centering
\resizebox{0.8\textwidth}{!}
{
\begin{tabular}{|c|c|c|c|c|c|c|c|} 
\hline
\multicolumn{2}{|c|}{Baselines} & \multicolumn{2}{c|}{2019 Competition Results} & \multicolumn{2}{c|}{2020 Competition Results} & \multicolumn{2}{c|}{2021 Competition Results}                         \\ 
\hline
Name    & Score                 & Team Name    & Score                          & Team Name           & Score                   & Team Name                                           & Score           \\ 
\hline
SQIL    & 2.94                  & CDS (ForgER) & 61.61                          & HelloWorld (SEIHAI) & 39.55                   & \textbf{X3 (JueWu-MC) }                             & \textbf{76.97}  \\
DQfD    & 2.39                  & mc\_rl       & 42.41                          & michal\_opanowicz   & 13.29                   & WinOrGoHome                                         & 22.97           \\
Rainbow & 0.42                  & I4DS         & 40.8                           & NoActionWasted      & 12.79                   & MCAgent                                             & 18.98           \\
PDDDQN  & 0.11                  & CraftRL      & 23.81                          & Rabbits             & 5.16                    & sneakysquids                                        & 14.35           \\
BC      & 2.40                  & UEFDRL       & 17.9                           & MajiManji           & 2.49                    & JBR\_HSE                                            & 10.33           \\
        &                       & TD240        & 15.19                          & BeepBoop            & 1.97                    & zhongguodui                                         & 8.84            \\
\hline
\end{tabular}
}
\caption{MineRL Competition Results. Our solution (JueWu-MC) significantly outperforms all other competitive solutions. }
\vspace{-0.1in}
\label{table:competition}
\end{table*}

\subsection{Learning Sub-policies to Craft Items}
In this type of sub-task, agents must learn a sequence of actions to craft items. For example, to craft a wooden pickaxe, agents need to craft planks/sticks/tables, place tables, and finally craft a wooden pickaxe.
In order to finish such tasks, agents need to learn a robust policy to execute a sequence of actions.

We explore pure imitation learning (IL) to reduce the need of interactions with the environment, due to the limited sample and interaction usage in MineRL. 
We propose \textit{ensemble behavior cloning with consistency filtering} (EBC). %

\paragraph{Consistency Filtering}
Human demonstrations can be diverse and noisy. Directly imitating such noisy data can cause confusion for the policy since there might appear different expert actions under the same state. In order words, there might exist multiple modes of near-optimal behavior as well as sub-optimal behaviors in human demonstrations, and it can be difficult for BC to capture consistent good behaviors by learning from human demonstrations. Therefore, we perform \textit{consistency filtering} by extracting the most common pattern of human behaviors. We show the algorithm in Appendix~\ref{app:algo}. We first extract the most common action sequence from demonstrations $\cD_i$ (line 1 to 10), and then keep those demonstrations that follow this action sequence while filtering the others, resulting in a smaller but consistent demonstration set $\bar{\cD_i}$ (line 11 to 13). %

\paragraph{Ensemble Behavior Cloning}
Learning policy from offline datasets can suffer from generalization issues. Policy learned by BC will become uncertain when encountering unseen out-of-distribution states, which leads to errorneous output at each step and cause compounding errors. 

To mitigate these issues, EBC learns a population of policies on different subsets of demonstrations to reduce the uncertainty of the agent's decision. Specifically, we train $K$ policies on different demonstrations with NLL loss:
\begin{equation} \label{eq:ebc}
\footnotesize
    \min_{\theta_k} \Esub{s,a \sim \bar{\cD_i}^k} \left[ - \log \pi_{\theta_k}(a|s) \right],   \bar{\cD}_i^k \subset \bar{\cD_i}, k=1,2,...,K,
\end{equation}
where $\theta_k$ parameterizes the $k$-th policy.
During inference, EBC adopts the majority voting mechanism to select an action that is the most confident among the population of policies. In this way, EBC learns a robust agent with reduced uncertainty and thus mitigates compounding errors.

%% file: exp.tex
\section{Experiment}

We conduct experiments on MineRL environment~\citep{MineRLDataset}. Our approach is built based on existing RL algorithms including SQIL~\citep{reddy2019sqil}, PPO~\citep{schulman2017proximal}, DQfD~\citep{hester2018deep}. 
We present the details of our approach as well as the experiment settings in Appendix~\ref{app:setting}.
We aim to answer the following questions: (1) How does our approach perform compared with baselines in terms of sample efficiency and final performance? (2) How does each proposed techniques contribute to the overall performance? (3) How do the proposed techniques work and help improve sample efficiency during training? %

\paragraph{Baselines}
We compare our approach with several baselines in two categories: 1)  end-to-end learning methods used in prior work and in the official implementation provided, including BC~\citep{kanervisto2020playing}, SQIL~\citep{reddy2019sqil}, Rainbow~\citep{hessel2018rainbow}, DQfD~\citep{hester2018deep}, PDDDQN~\citep{schaul2015prioritized,van2016deep,wang2016dueling}; all these baselines are trained within 8 million samples with default hyper-parameters. 2) The top solutions from MineRL 2019\&2020\&2021 including SEIHAI~\citep{mao2021seihai} (1st place in MineRL 2020) and ForgER~\citep{skrynnik2021forgetful} (1st place in MineRL 2019). %

\begin{table}
\centering
\resizebox{0.46\textwidth}{!}
{
\begin{tabular}{|c|c|c|c|c|c|c|c|} 
\hline
\diagbox{Methods}{Stage} & 1 & 2  & 3  & 4  & 5 & 6  & 7  \\ 
\hline
SEIHAI~\citep{mao2021seihai}                   & 64\%          & 78.6\%            & 78.3\%          & 84.7\%            & 23\%               & 0\%               & 0\%                \\
JueWu-MC                 & \textbf{92\%}          & \textbf{96\%}              & \textbf{96\%}            & \textbf{87\%}              & \textbf{46\%}               & \textbf{11\%}               & 0\%                \\
\hline
\end{tabular}
}
\caption{The conditional success rate of each stage.}%
\vspace{-0.15in}
\label{table:passrate}
\end{table}

\subsection{Main Results}
Table~\ref{table:competition} shows all the MineRL competition results since it was held in 2019. 
It is important to mention that the competition settings in MineRL 2020\&2021 are much more difficult than MineRL 2019, e.g, the obfuscation in state and action space, making it impossible to work on action/state tricks. In this case, participants have to focus on the algorithm design itself in 2020 and 2021. 
Therefore, the scores in MineRL 2020\&2021 are lower than those in MineRL 2019, which is explained in \citep{guss2021minerl}. 
Overall, our approach outperforms all previous solutions, even including MineRL 2019, by a very large margin. 
We also find that end-to-end baselines can not get a decent result, which indicates that it is quite challenging to solve such a long-horizon task with end-to-end learning. Compared with the results of MineRL 2020 (directly comparable with MineRL 2021 as they share the same settings), our method significantly outperforms other competitive solutions with a score (76.97) that is 3.4x higher compared to the second place (22.97), and even outperforms the sum of all other competitors' scores (75.25) in MineRL 2020. 
We also compare the conditional success rate of each stage between our approach and SEIHAI in Table~\ref{table:passrate} and find that our method outperforms SEIHAI in every stages clearly. 

\begin{figure}[t]
	\centering
	\vspace{-0.2in}
    \subfigure[]{
        \label{fig:curve}
        \includegraphics[width=0.2\textwidth]{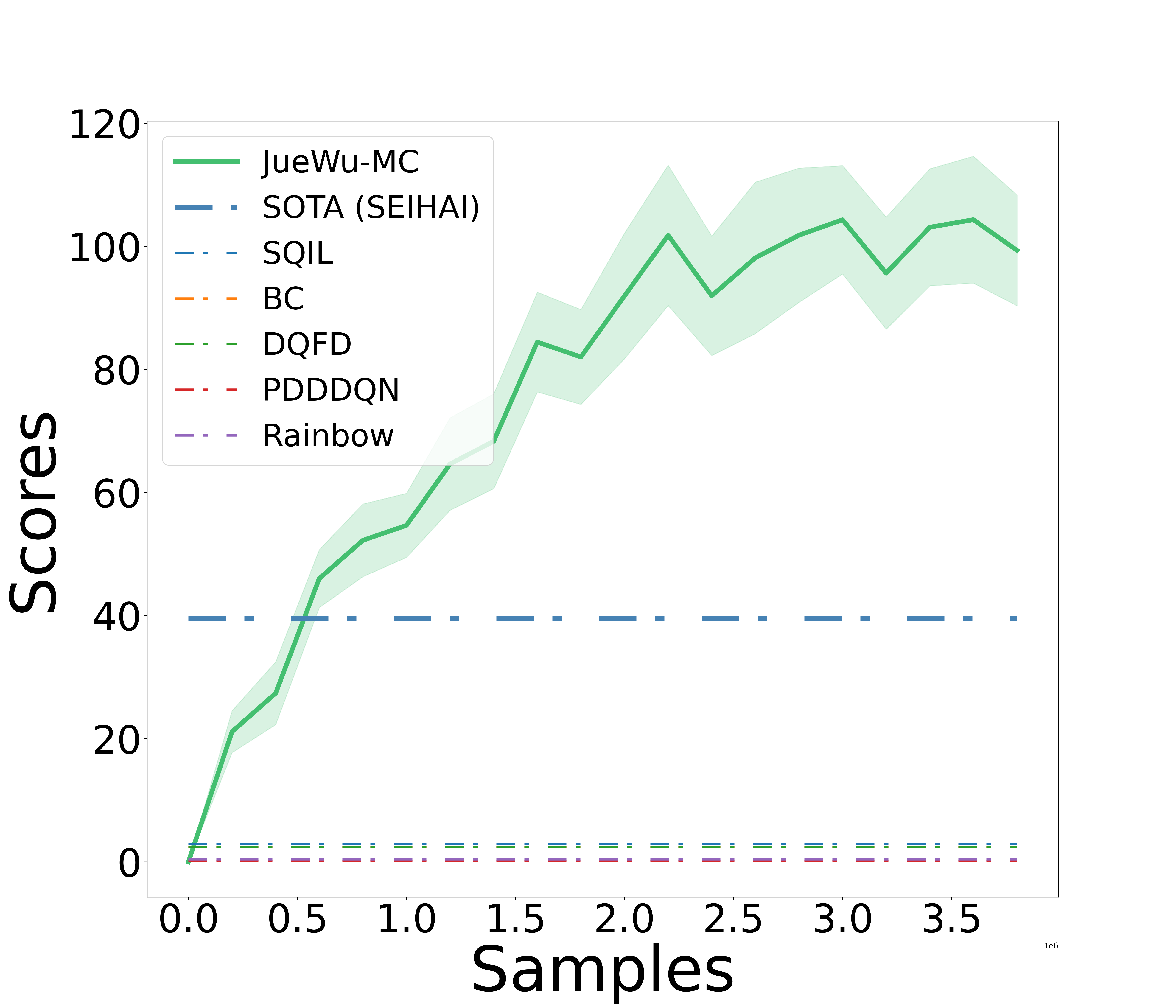}
    }
	\subfigure[]{
        \label{fig:ablation}
	    \includegraphics[width=0.2\textwidth]{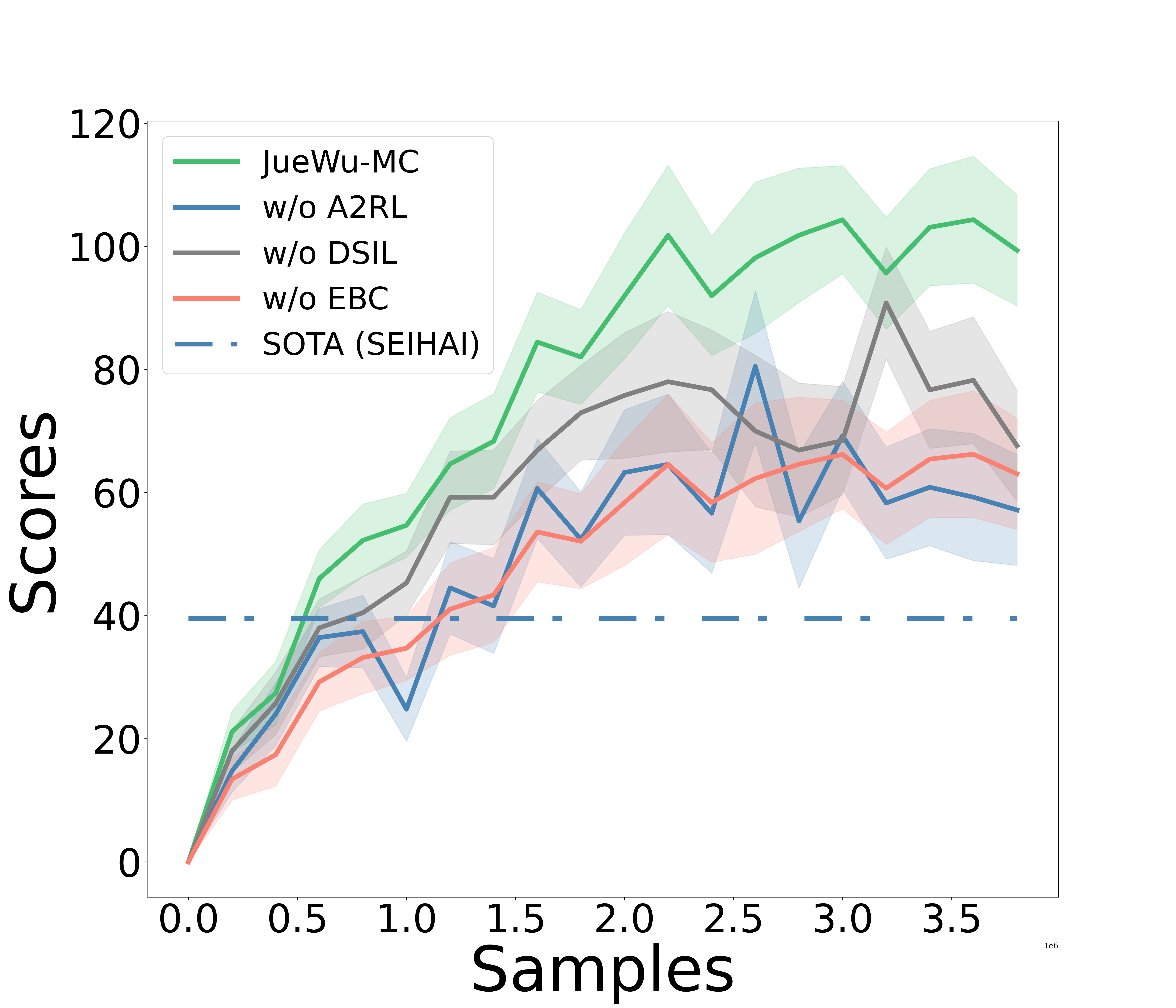}
    }
    \vspace{-0.1in}
	\caption{(a) The evaluation results along with training process. (b) Ablation study of our propose techniques.}
	\vspace{-0.1in}
\end{figure}

We show the training curves in Figure~\ref{fig:curve}. Note that due to the version update of MineRL 2021 (0.3.7 $\rightarrow$ 0.4.2) in online platform, many teams suffer from a performance drop during online evaluation. Therefore, our online score (76.97) has a drop comparing with the performance in our training curve (>100). Even though, all the baselines we compare in the training curves share the same version (0.3.7) of MineRL, which is a fair comparison.
Our approach is sample-efficient and outperforms the prior best results with only 0.5 million training samples. Our score reaches 100 with only 2.5 million training samples, which is much less than the limited number of samples (8 million) in MineRL competition. 
We do not include the training curve of ForgER~\citep{skrynnik2021forgetful} because their environment setting (MineRL 2019) is easier than ours and thus is not comparable in sample efficiency.

\subsection{Ablation Study}
To examine the effectiveness of our proposed techniques, we consider three variants of our approach: 1) w/o A2RL, 2) w/o DSIL, and 3) w/o EBC. We show the training curves for each in Figure~\ref{fig:ablation}. We find that each proposed technique has a decent contribution to the overall performance. 
We also find that both EBC and A2RL contribute more than DSIL. This is because DSIL mainly boosts the performance for the later sub-tasks (i.e., stage 5 in appendix) while A2RL and EBC have earlier effects to the overall pipeline. %
EBC also contributes a lot to the overall performance, which demonstrates that learning a robust policy is quite important to solve a long-horizon task.

\subsection{Visualization}
\begin{figure}[t]
	\centering
	\vspace{-0.1in}
	\includegraphics[width=0.43\textwidth]{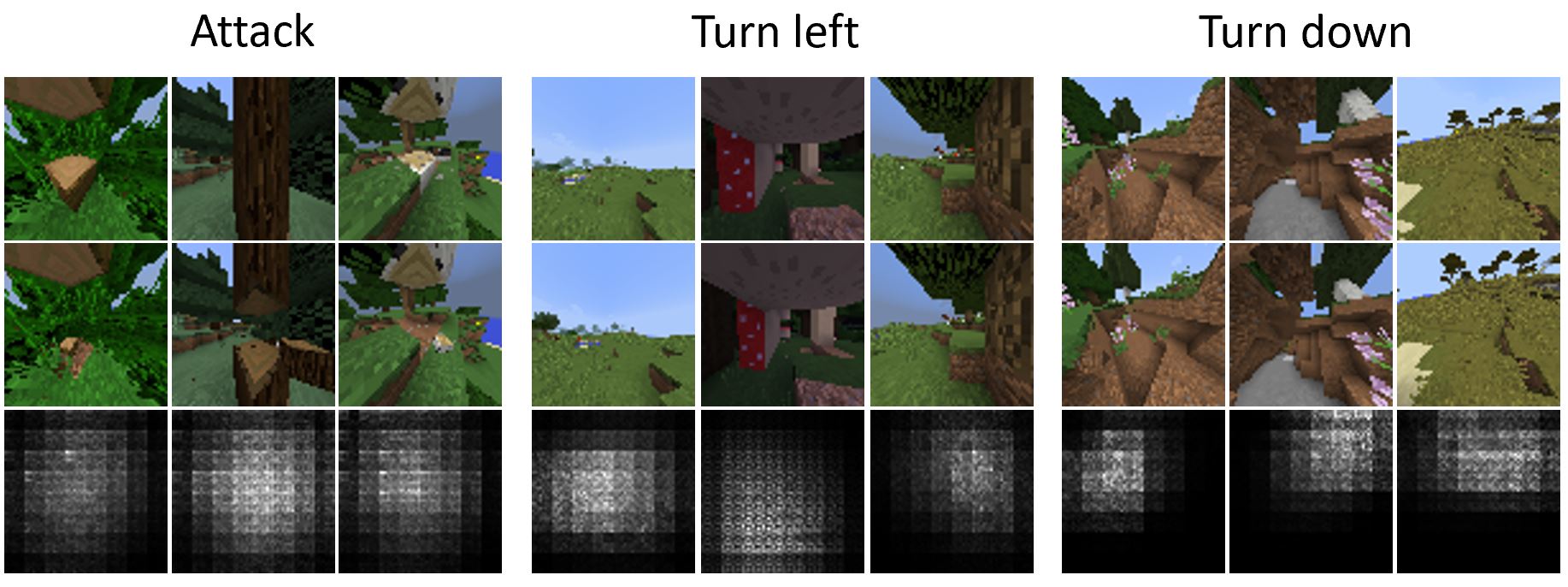}
	\caption{Saliency map for the learned mask. %
	Top row: current state; middle: next state; bottom: saliency map of the mask.}
	\vspace{-0.1in}
	\label{fig:saliency}
\end{figure}

To gain insights into why our learning techniques work, we conduct an in-depth analysis. First, to understand the learned mask in A2RL, we compute saliency maps~\citep{simonyan2013deep} for the action-aware mask generator. Specifically, to visualize the salient part of the images as seen by the mask net, we compute the absolute value of the Jacobian $|\nabla_s m_{\phi}(s,a)|$. Figure~\ref{fig:saliency} shows the visualizations for three actions (attack, turn left, turn down). For each action $a$, we show current states, next states, and the saliency map of the learned mask on current states.
We find that the learned mask is able to capture the dynamic information between two adjacent states, revealing the curiosity on the effect of actions. 
Specifically, the mask net learns to focus on uncertain parts of the current state, i.e., the parts that introduce uncertainty to the dynamic model. By focusing on the uncertain parts, the mask net learns to provide multiple views for the current state which can reflect the effects of different actions. 
For the `attack' action, the learned mask attends to objects in front of the agent --- this can help the agent pay more attention to the objects that will be attacked. For the `turn left' and `turn down' actions, the mask net learns to focus on the parts that have major change due to the rotation and translation of the agent's perspective. 
Our learned mask assists the agent to better understand the 3D environment, which thus enables the agent to learn to control efficiently. Moreover, A2RL improves the interpretability of the learned policy.

\begin{figure}[t]
    \centering
    \subfigure[PPO early] 
    {
        \includegraphics[width=.1\textwidth]{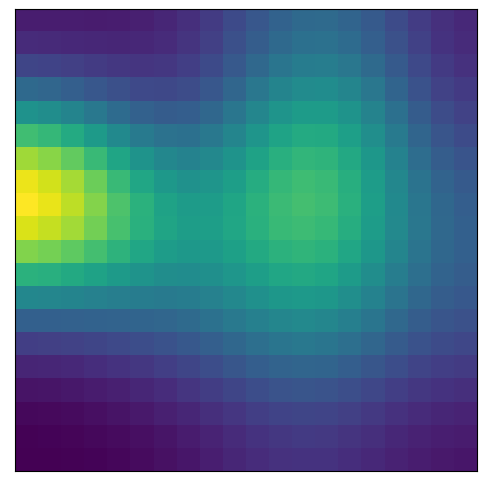}
    }
    \subfigure[SIL early] 
    {
        \includegraphics[width=.1\textwidth]{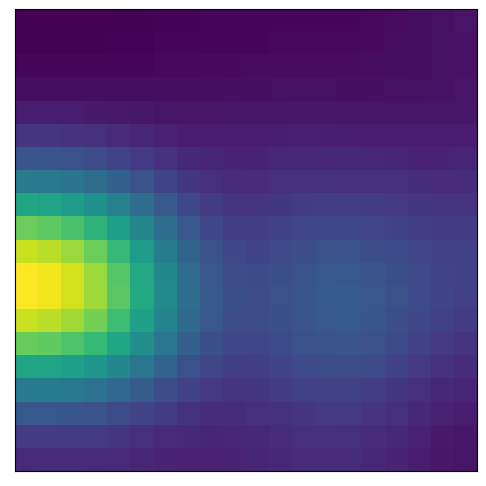}
    }
    \subfigure[DSIL early] 
    {
        \includegraphics[width=.1\textwidth]{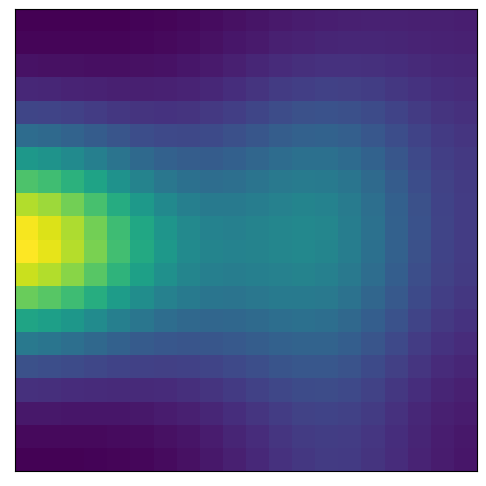}
    }
    \\
    \vspace{-0.1in}
    \subfigure[PPO late] 
    {
        \includegraphics[width=.1\textwidth]{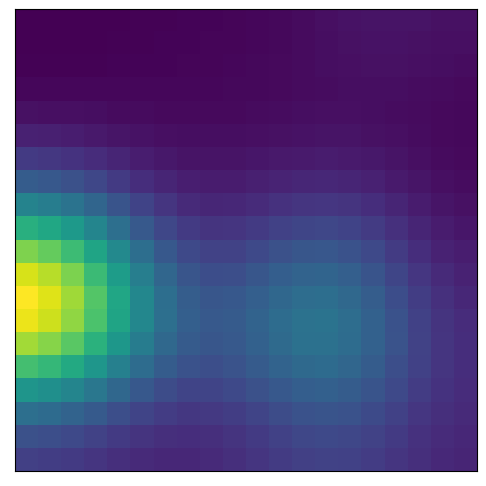}
    }
    \subfigure[SIL late] 
    {
        \includegraphics[width=.1\textwidth]{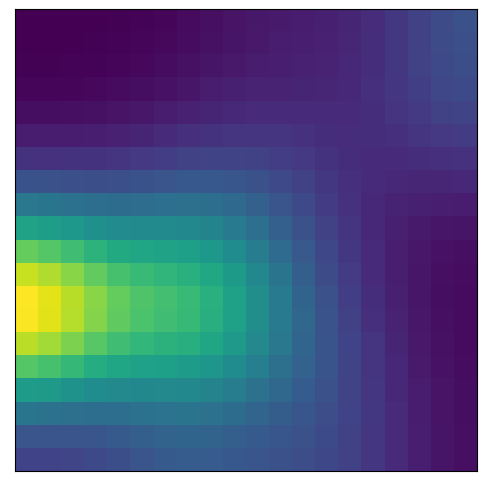}
    }
    \subfigure[DSIL late] 
    {
        \includegraphics[width=.1\textwidth]{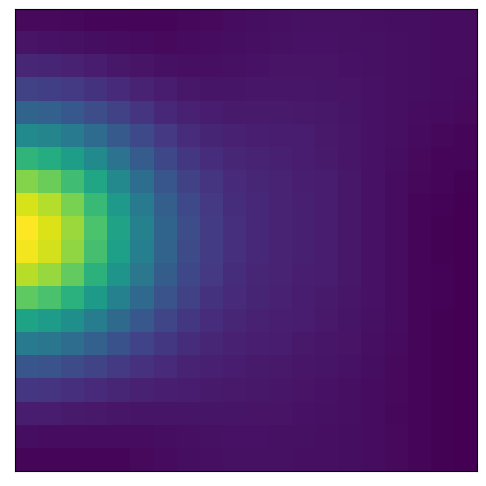}
    }
    \vspace{-0.1in}
    \caption{Visualization of visited state distribution during training.}
    \label{fig:tsne}
\end{figure}

To understand how DSIL works, we also visualize the state distribution that the agent visits. 
We compare `PPO', `PPO+SIL' and `PPO+DSIL'. For each method, we plot the visiting state distribution in different training stages. Figure~\ref{fig:tsne} shows the visualization results. At the early training stage, both methods explore randomly and sometimes get to the goal state successfully. After getting enough samples and training for several epochs, `PPO+DSIL' starts to explore in a more compact region, %
while `PPO' and `PPO+SIL' still explore in a wide region. %
This is because DSIL can push the agent to stay close to good state distribution, reproduce its past good behaviors and explore in the right way, which incentivizes the agent's deep exploration to get successful trajectories.

%% file: conclusion.tex
\section{Conclusion}
In this paper, we present JueWu-MC, a sample-efficient hierarchical reinforcement learning framework designed to play Minecraft. With a high-level controller and several auto-extracted low-level workers, our framework can adapt to different environments and solve a series of sophisticated tasks. Furthermore, by introducing novel techniques on representation learning and imitation learning, we improve the performance and learning efficiency of the sub-policies. Experiments show that our pipeline significantly outperforms all the baseline algorithms and the solutions from past MineRL competitions. In future work, we would like to apply JueWu-MC to other Minecraft tasks, as well as other open-world games.

%% file: appendix.tex
\section*{Appendix}

\input{prelim}

\subsection*{Experiment Setting}\label{app:setting}
\paragraph{Training Data}
We use MineRL Diamond dataset [\cite{MineRLDataset}] which consists of 60 million samples (state/action/reward) across a variety of related tasks in Minecraft. Each state consists of an image with $64 \times 64$ resolutions and an obfuscated 64-dimensional inventory vector. Actions also have been obfuscated to 64-dimensional vectors.

\paragraph{Models}
We describe the models that we use in our hierarchical framework --- meta-policy: we use neural networks with 3-layer MLPs and ReLU activation; stage 1 (Treechop): we use SQIL~\citep{reddy2019sqil}, which learns from demonstrations with a constant reward function, and we also plug our A2RL technique into SQIL to boost perception; stage 2 (WoodAxe): we use our proposed EBC that learns policies by purely imitating from human demonstrations; stage 3 (Stone): we use DQfD~\citep{hester2018deep} but only use it to pre-train our sub-policy with demonstrations and do not interact with the environment; stage 4 (Furnace): we use random search method that randomly samples action in the action space; stage 5 (Iron Ore): we use PPO~\citep{schulman2017proximal} with our proposed A2RL and DSIL; stage 6 (IronAxe): we use random search method as in stage 4; stage 7 (Diamond): we use the same training procedure as used in stage 5. Note that the action space in each stage is automatically extracted from the structured demonstrations. We describe the details as well as hyper-parameters of each stage below. All models are optimized with Adam.  %

\begin{itemize}[leftmargin=*]
    \item Stage 1: we use SQIL with A2RL in this stage. The number of action clusters are set to 64. For A2RL, we set $\eta=0.1$. To guide the agent in the sparse-reward environment, we do reward shaping to boost sample efficiency. Specifically, we modified the constant reward function of SQIL with explicitly guided reward shaping. We shape the reward function so that the agent gets increasing rewards when it approaches the rewarding states. This helps our agent achieve a decent performance at the very beginning. %
    \item Stage 2: we use EBC to learn $K=10$ policies simultaneously with a learning rate 1e-4. We set the number of clusters in KMeans as 5.
    \item Stage 3: we use the default hyper-parameter setting of the original DQfD and set the number of action clusters as 10.
    \item Stage 4: we set the number of action clusters as 3. Since the action space is small, we find that the performance of a simple random search method is already comparable to EBC. Therefore, we adopt random search in this stage.
    \item Stage 5: we use default settings of PPO, and integrate A2RL and DSIL into PPO. For A2RL, we set $\eta=0.1$ and set the learning rate of mask net as 1e-4. For DSIL, we set $\epsilon=0.1$ and the coefficient of intrinsic reward as 0.01. The learning rate for training the discriminator is set to 1e-4.
    \item Stage 6: the number of action clusters here is the same as stage 4 (which has 3 actions). Therefore we adopt the simple random search again and find that such a simple method can already achieve decent performance.
    \item Stage 7: for the final stage we use the same approach as in stage 5 to learn an individual policy. %
\end{itemize}

\paragraph{Evaluation}
We train our approach for 4 million samples, and evaluate the overall performance every 0.2 million. For each evaluation, we roll out 100 episodes with different random seeds and take the average of 100 scores. We also evaluate our approach with three sets of random seeds to examine the variance of results.

\subsection*{Algorithm}\label{app:algo}
Algorithm~\ref{algo:A2RL_DSIL} shows how to learn sub-policies to gather resources with our proposed A2RL and DSIL techniques. We first initialize two buffers to store successful trajectories and failed trajectories. Our agent interacts with the environment to collect samples (line 3 to 7) and stores samples to the buffers accordingly (line 8). Before training the mask generator, we train the projection net (line 9 to 11). Then, we train the mask net (line 12 to 14) and the discriminator (line 15 to 17) accordingly. Finally, we perform policy updates with off-the-shelf RL algorithms (line 18 to 20).

Algorithm~\ref{algo:EBC} shows how to learn sub-policies to craft items with our proposed EBC technique. Note that EBC does not need to interact with the environment. Given human demonstrations, our approach first extracts the most common action pattern (line 3 to 9). Specifically, for each trajectory, we keep those actions that lead to state change while appearing at the first time (line 6) to form an action sequence, and count the occurrences of each action pattern (line 8 to 9). Then, we get the most common action pattern (line 10). Afterward, we conduct consistency filtering using the extracted action pattern (line 11 to 13). Finally, we learn a population of policies using ensemble behavior cloning (line 14 to 16).

\begin{algorithm}[h]%
\caption{Learning to Gather Resources with Action-aware Representation and Discriminator-based Self-Imitation}
\label{algo:A2RL_DSIL}
\begin{algorithmic}[1]
    \State Initialize success buffer $\cB_{i}^{+}$ and failure buffer $\cB_{i}^{-}$
    \For {Episode number $e=1...E$}
        \State Receive initial observation $s_0$ from environment.
        \For {steps $t \in {0,1,2,...,k}$}
            \State Sample action $a_t \sim \pi(a|s)$
            \State Take action $a_t$, receive reward $r_t$ and next state $s_{t+1}$
            \State Store $(s_t, a_t, r_t, s_{t+1})$ to $\cD$
        \EndFor
        \State If episode succeeds, append all transitions $(s,a,s')$ in trajectory $\tau_e$ to $\cB_{i}^{+}$, otherwise, append $(s,a,s')$ to $\cB_{i}^{-}$.
        \While {$g_{\psi_a}$ not converge}  \Comment{Training projection net}
            \State Sample training batch $(s,a,s')$ from $\cD$
            \State Update with Eq~\ref{eq:g}: $\psi_a \leftarrow \psi_a - \alpha \nabla_{\psi_a} \cL_{g}(\psi_a)$
        \EndWhile
        \While {$m_{\phi}$ not converge}  \Comment{Training mask net}
            \State Sample training batch $(s,a,s')$ from $\cD$
            \State Update with Eq~\ref{eq:mask}: $\phi \leftarrow \phi - \beta \nabla_{\phi} \cL_{m}(\phi)$
        \EndWhile
        \While {$D_{\xi}$ not converge}  \Comment{Training discriminator}
            \State Sample positive data from $\cB_{i}^{+}$ and negative data from $\cB_{i}^{-}$.
            \State Update discriminator with Eq~\ref{eq:discriminator}.
        \EndWhile
        \For {$K_p$ training iterations}  \Comment{Policy Iteration}
            \State Compute intrinsic reward with Eq~\ref{eq:intrinsic}.
            \State Perform policy update using any RL algorithms. (The parameters of feature maps $f_{\theta}(s)$ in Eq.~\ref{eq:mask_L1}, \ref{eq:mask_L2}, \ref{eq:g} are optimized with RL objectives.)
        \EndFor
    \EndFor
\end{algorithmic}
\end{algorithm}

\begin{algorithm}[h] %
\caption{Learning to Craft Items using Ensemble Behavior Cloning with Consistency Filtering}
\label{algo:EBC}
\begin{algorithmic}[1]
    \State Initialize empty buffer $\bar{\cD_i}$.
    \State Initialize pattern counter $c$, pattern recorder $p$.
    \For {Trajectory $\tau_e \in \cD_i$}  \Comment{Get action patterns}
        \State $\text{seq} = []$
        \For {each transition $(s,a,s') \in \tau_e$}
            \If {$s \neq s'$ and $a \notin \text{seq}$}
                \State $\text{seq}$.append($a$)
            \EndIf
        \EndFor
        \State Update the counter: $c[\text{seq}]$ = $c[\text{seq}]$ + 1.
        \State Record the pattern: $p[\tau_e]$ = $\text{seq}$.
    \EndFor
    \State Get the most common pattern $u = \argmax_{\text{seq}} c[\text{seq}]$
    \For {Trajectory $\tau_e \in \cD_i$}  \Comment{Consistency filtering}
        \If {$p[\tau_e]$ == $u$}
            \State $\bar{\cD_i}$.append($\tau_e$)
        \EndIf
    \EndFor
    \State Randomly split $\bar{\cD_i}$ into $K$ subsets $\bar{\cD}_i^k \subset \bar{\cD_i}, k=1,2,...,K$.
    \For {$k=1,2,...,K$} \Comment{Ensemble behavior cloning}
        \State Train the $k$-th policy using demonstration $\bar{\cD}_i^k$ with Eq~\ref{eq:ebc}.
    \EndFor
\end{algorithmic}
\end{algorithm}

%% file: prelim.tex
\subsection*{Minecraft \& MineRL}

Minecraft is a 3D open-world video game which has been extensively used in education and research \citep{nebel2016mining}. 
Recently, it also becomes a popular testbed for RL research \citep{oh2017zero,oh2016control,guss2021minerl}, due to its playing mechanism (centered around exploration in open-world) and similarity to real-world tasks (perception and construction). 
MineRL \citep{MineRLDataset}, since 2019, is an annual research competition focusing on the development of sample efficient RL algorithms for mining items in Minecraft. 

\begin{wraptable}{R}{.25\textwidth} %
\centering
\scriptsize
\vspace{-0.1in}
\begin{tabular}{l|l|l|l}
\textbf{Item} & \textbf{r} & \textbf{Item} & \textbf{r} \\ \hline
Log           & 1               & StoneAxe      & 32              \\
Planks        & 2               & Furnace       & 32              \\
Sticks        & 4               & IronOre       & 64              \\
CraftTable    & 4               & IronIngot     & 128             \\
WoodAxe       & 8               & IronAxe       & 256             \\
Stone         & 16              & Diamond*      & 1024           
\end{tabular}
\caption{Reward in MineRL.}
\label{table:reward}
\vspace{-0.1in}
\end{wraptable}

In MineRL, state space $\cS$ can be represented as $\{ \tilde{\cS}, \bar{\cS} \}$ where $\tilde{s} \in \tilde{\cS}$ is of 64x64 pixel and $\bar{s} \in \bar{\cS}$ is an inventory observation representing obtained items. 
The action space $\cA$ is the Cartesian product of view adjustment (continuous), move (discrete), and actions for placing blocks, crafting items, smelting items, and mining/hitting enemies (discrete). The reward $r$ is highly sparse, where only the first time of obtaining each item on diamond's prerequisite chain is accordingly rewarded (refer to Table~\ref{table:reward}). In MineRL 2021, the action space and inventory space are obfuscated, which prevents competitors from hand-crafting actions or rule-based algorithms.

%% file: main.bbl
\begin{thebibliography}{}

\bibitem[\protect\citeauthoryear{Amiranashvili \bgroup \em et al.\egroup
  }{2020}]{amiranashvili2020scaling}
Artemij Amiranashvili, Nicolai Dorka, et~al.
\newblock Scaling imitation learning in minecraft.
\newblock {\em arXiv preprint arXiv:2007.02701}, 2020.

\bibitem[\protect\citeauthoryear{Berner \bgroup \em et al.\egroup
  }{2019}]{berner2019dota}
Christopher Berner, Greg Brockman, Brooke Chan, et~al.
\newblock Dota 2 with large scale deep reinforcement learning.
\newblock {\em arXiv preprint arXiv:1912.06680}, 2019.

\bibitem[\protect\citeauthoryear{Burda \bgroup \em et al.\egroup
  }{2018}]{burda2018exploration}
Yuri Burda, Harrison Edwards, et~al.
\newblock Exploration by random network distillation.
\newblock {\em arXiv preprint arXiv:1810.12894}, 2018.

\bibitem[\protect\citeauthoryear{Guss \bgroup \em et al.\egroup
  }{2019}]{MineRLDataset}
William~H. Guss, Brandon Houghton, et~al.
\newblock Minerl: {A} large-scale dataset of minecraft demonstrations.
\newblock In {\em IJCAI}, 2019.

\bibitem[\protect\citeauthoryear{Guss \bgroup \em et al.\egroup
  }{2021}]{guss2021minerl}
William~H Guss, Mario~Ynocente Castro, et~al.
\newblock The minerl 2020 competition on sample efficient reinforcement
  learning using human priors.
\newblock {\em arXiv:2101.11071}, 2021.

\bibitem[\protect\citeauthoryear{Hessel \bgroup \em et al.\egroup
  }{2018}]{hessel2018rainbow}
Matteo Hessel, Joseph Modayil, Hado Van~Hasselt, Tom Schaul, Georg Ostrovski,
  Will Dabney, Dan Horgan, Bilal Piot, Mohammad Azar, and David Silver.
\newblock Rainbow: Combining improvements in deep reinforcement learning.
\newblock In {\em AAAI}, 2018.

\bibitem[\protect\citeauthoryear{Hester \bgroup \em et al.\egroup
  }{2018}]{hester2018deep}
Todd Hester, Matej Vecerik, et~al.
\newblock Deep q-learning from demonstrations.
\newblock In {\em AAAI}, 2018.

\bibitem[\protect\citeauthoryear{Huang \bgroup \em et al.\egroup
  }{2019}]{huang2019combo}
Shiyu Huang, Hang Su, et~al.
\newblock Combo-action: Training agent for fps game with auxiliary tasks.
\newblock In {\em AAAI}, 2019.

\bibitem[\protect\citeauthoryear{Kanervisto \bgroup \em et al.\egroup
  }{2020}]{kanervisto2020playing}
Anssi Kanervisto, Janne Karttunen, and Ville Hautam{\"a}ki.
\newblock Playing minecraft with behavioural cloning.
\newblock In {\em NeurIPS 2019 Competition and Demonstration Track}, 2020.

\bibitem[\protect\citeauthoryear{Krishna and Murty}{1999}]{krishna1999genetic}
K~Krishna and M~Narasimha Murty.
\newblock Genetic k-means algorithm.
\newblock {\em IEEE Trans. on Systems, Man and Cyber.}, 1999.

\bibitem[\protect\citeauthoryear{Kurach \bgroup \em et al.\egroup
  }{2020}]{kurach2020google}
Karol Kurach, Anton Raichuk, Piotr Sta{\'n}czyk, et~al.
\newblock Google research football: A novel reinforcement learning environment.
\newblock In {\em AAAI}, 2020.

\bibitem[\protect\citeauthoryear{Le \bgroup \em et al.\egroup
  }{2018}]{le2018hierarchical}
Hoang Le, Nan Jiang, et~al.
\newblock Hierarchical imitation and reinforcement learning.
\newblock In {\em ICML}, 2018.

\bibitem[\protect\citeauthoryear{Lesort \bgroup \em et al.\egroup
  }{2018}]{lesort2018state}
Timoth{\'e}e Lesort, Natalia D{\'\i}az-Rodr{\'\i}guez, Jean-Franois Goudou, and
  David Filliat.
\newblock State representation learning for control: An overview.
\newblock {\em Neural Networks}, 2018.

\bibitem[\protect\citeauthoryear{Mao \bgroup \em et al.\egroup
  }{2021}]{mao2021seihai}
Hangyu Mao, Chao Wang, et~al.
\newblock Seihai: A sample-efficient hierarchical ai for the minerl
  competition.
\newblock {\em arXiv preprint arXiv:2111.08857}, 2021.

\bibitem[\protect\citeauthoryear{Mnih \bgroup \em et al.\egroup
  }{2013}]{mnih2013playing}
Volodymyr Mnih, Koray Kavukcuoglu, et~al.
\newblock Playing atari with deep reinforcement learning.
\newblock {\em arXiv preprint arXiv:1312.5602}, 2013.

\bibitem[\protect\citeauthoryear{Nair \bgroup \em et al.\egroup
  }{2018}]{nair2018overcoming}
Ashvin Nair, Bob McGrew, Marcin Andrychowicz, Wojciech Zaremba, and Pieter
  Abbeel.
\newblock Overcoming exploration in reinforcement learning with demonstrations.
\newblock In {\em ICRA}, 2018.

\bibitem[\protect\citeauthoryear{Nebel \bgroup \em et al.\egroup
  }{2016}]{nebel2016mining}
Steve Nebel, Sascha Schneider, et~al.
\newblock Mining learning and crafting scientific experiments: a literature
  review on the use of minecraft in education and research.
\newblock 19(2):355--366, 2016.

\bibitem[\protect\citeauthoryear{Oh \bgroup \em et al.\egroup
  }{2016}]{oh2016control}
Junhyuk Oh, Valliappa Chockalingam, Honglak Lee, et~al.
\newblock Control of memory, active perception, and action in minecraft.
\newblock In {\em ICML}, 2016.

\bibitem[\protect\citeauthoryear{Oh \bgroup \em et al.\egroup
  }{2017}]{oh2017zero}
Junhyuk Oh, Satinder Singh, Honglak Lee, and Pushmeet Kohli.
\newblock Zero-shot task generalization with multi-task deep reinforcement
  learning.
\newblock In {\em International Conference on Machine Learning}, pages
  2661--2670. PMLR, 2017.

\bibitem[\protect\citeauthoryear{Oh \bgroup \em et al.\egroup
  }{2018}]{oh2018self}
Junhyuk Oh, Yijie Guo, Satinder Singh, and Honglak Lee.
\newblock Self-imitation learning.
\newblock In {\em ICML}, 2018.

\bibitem[\protect\citeauthoryear{Pathak \bgroup \em et al.\egroup
  }{2017}]{pathak2017curiosity}
Deepak Pathak, Pulkit Agrawal, Alexei~A Efros, and Trevor Darrell.
\newblock Curiosity-driven exploration by self-supervised prediction.
\newblock In {\em ICML}, 2017.

\bibitem[\protect\citeauthoryear{Pertsch \bgroup \em et al.\egroup
  }{2020}]{pertsch2020accelerating}
Karl Pertsch, Youngwoon Lee, and Joseph~J Lim.
\newblock Accelerating reinforcement learning with learned skill priors.
\newblock {\em arXiv preprint arXiv:2010.11944}, 2020.

\bibitem[\protect\citeauthoryear{Reddy \bgroup \em et al.\egroup
  }{2019}]{reddy2019sqil}
Siddharth Reddy, Anca~D Dragan, and Sergey Levine.
\newblock Sqil: Imitation learning via reinforcement learning with sparse
  rewards.
\newblock {\em arXiv preprint arXiv:1905.11108}, 2019.

\bibitem[\protect\citeauthoryear{Schaul \bgroup \em et al.\egroup
  }{2015}]{schaul2015prioritized}
Tom Schaul, John Quan, Ioannis Antonoglou, and David Silver.
\newblock Prioritized experience replay.
\newblock {\em arXiv preprint arXiv:1511.05952}, 2015.

\bibitem[\protect\citeauthoryear{Scheller \bgroup \em et al.\egroup
  }{2020}]{scheller2020sample}
Christian Scheller, Yanick Schraner, and Manfred Vogel.
\newblock Sample efficient reinforcement learning through learning from
  demonstrations in minecraft.
\newblock In {\em NeurIPS 2019 Competition and Demonstration Track}, pages
  67--76. PMLR, 2020.

\bibitem[\protect\citeauthoryear{Schulman \bgroup \em et al.\egroup
  }{2017}]{schulman2017proximal}
John Schulman, Filip Wolski, et~al.
\newblock Proximal policy optimization algorithms.
\newblock {\em arXiv preprint arXiv:1707.06347}, 2017.

\bibitem[\protect\citeauthoryear{Silver \bgroup \em et al.\egroup
  }{2016}]{silver2016mastering}
David Silver, Aja Huang, et~al.
\newblock Mastering the game of go with deep neural networks and tree search.
\newblock {\em nature}, 2016.

\bibitem[\protect\citeauthoryear{Simonyan \bgroup \em et al.\egroup
  }{2013}]{simonyan2013deep}
Karen Simonyan, Andrea Vedaldi, and Andrew Zisserman.
\newblock Deep inside convolutional networks: Visualising image classification
  models and saliency maps.
\newblock {\em arXiv preprint arXiv:1312.6034}, 2013.

\bibitem[\protect\citeauthoryear{Skrynnik \bgroup \em et al.\egroup
  }{2021}]{skrynnik2021forgetful}
Alexey Skrynnik, Aleksey Staroverov, et~al.
\newblock Forgetful experience replay in hierarchical reinforcement learning
  from expert demonstrations.
\newblock {\em Knowledge-Based Systems}, 2021.

\bibitem[\protect\citeauthoryear{Srinivas \bgroup \em et al.\egroup
  }{2020}]{srinivas2020curl}
Aravind Srinivas, Michael Laskin, and Pieter Abbeel.
\newblock Curl: Contrastive unsupervised representations for reinforcement
  learning.
\newblock {\em arXiv preprint arXiv:2004.04136}, 2020.

\bibitem[\protect\citeauthoryear{Tessler \bgroup \em et al.\egroup
  }{2017}]{tessler2017deep}
Chen Tessler, Shahar Givony, et~al.
\newblock A deep hierarchical approach to lifelong learning in minecraft.
\newblock In {\em AAAI}, 2017.

\bibitem[\protect\citeauthoryear{Van~Hasselt \bgroup \em et al.\egroup
  }{2016}]{van2016deep}
Hado Van~Hasselt, Arthur Guez, et~al.
\newblock Deep reinforcement learning with double q-learning.
\newblock In {\em AAAI}, 2016.

\bibitem[\protect\citeauthoryear{Vinyals \bgroup \em et al.\egroup
  }{2019}]{vinyals2019grandmaster}
Oriol Vinyals, Igor Babuschkin, et~al.
\newblock Grandmaster level in starcraft ii using multi-agent reinforcement
  learning.
\newblock {\em Nature}, 575(7782):350--354, 2019.

\bibitem[\protect\citeauthoryear{Wang \bgroup \em et al.\egroup
  }{2016}]{wang2016dueling}
Ziyu Wang, Tom Schaul, Matteo Hessel, Hado Hasselt, Marc Lanctot, and Nando
  Freitas.
\newblock Dueling network architectures for deep reinforcement learning.
\newblock In {\em ICML}, 2016.

\bibitem[\protect\citeauthoryear{Whitney \bgroup \em et al.\egroup
  }{2019}]{whitney2019dynamics}
William Whitney, Rajat Agarwal, Kyunghyun Cho, and Abhinav Gupta.
\newblock Dynamics-aware embeddings.
\newblock {\em arXiv preprint arXiv:1908.09357}, 2019.

\bibitem[\protect\citeauthoryear{Wu \bgroup \em et al.\egroup
  }{2019}]{wu2019imitation}
Yueh-Hua Wu, Nontawat Charoenphakdee, et~al.
\newblock Imitation learning from imperfect demonstration.
\newblock In {\em ICML}, 2019.

\bibitem[\protect\citeauthoryear{Wu \bgroup \em et al.\egroup
  }{2021}]{wu2021self}
Haiping Wu, Khimya Khetarpal, and Doina Precup.
\newblock Self-supervised attention-aware reinforcement learning.
\newblock In {\em AAAI}, 2021.

\end{thebibliography}
